\documentclass[10pt,twocolumn,letterpaper]{article}

\PassOptionsToPackage{dvipsnames,table}{xcolor}
\usepackage[pagenumbers]{iccv} 

\usepackage{latexsym}
\usepackage{amssymb}
\usepackage{amsmath}
\usepackage{amsthm}
\usepackage{booktabs}
\usepackage{enumitem}
\usepackage{graphicx}
\usepackage{color}
\usepackage{booktabs}
\usepackage{url}

\newcommand{\BibTeX}{B\kern-.05em{\sc i\kern-.025em b}\kern-.08em\TeX}

\definecolor{iccvblue}{rgb}{0.21,0.49,0.74}
\usepackage[pagebackref,breaklinks,colorlinks,allcolors=iccvblue]{hyperref}

\def\paperID{0016} 
\def\confName{ICCV}
\def\confYear{2025}

\title{NoiseCutMix: A Novel Data Augmentation Approach by Mixing Estimated Noise in Diffusion Models}

\author{
Shumpei Takezaki\thanks{Equal contribution}
\and
Ryoma Bise
\and
Shinnosuke Matsuo\footnotemark[1]
\and \
{Kyushu University, Fukuoka, Japan}
\and
{\tt\small \{shumpei.takezaki,shinnosuke.matsuo\}@human.ait.kyushu-u.ac.jp}
}

\begin{document}
\maketitle

\begin{abstract}
In this study, we propose a novel data augmentation method that introduces the concept of CutMix into the generation process of diffusion models, thereby exploiting both the ability of diffusion models to generate natural and high-resolution images and the characteristic of CutMix, which combines features from two classes to create diverse augmented data. 
Representative data augmentation methods for combining images from multiple classes include CutMix and MixUp. However, techniques like CutMix often result in unnatural boundaries between the two images due to contextual differences.
Therefore, in this study, we propose a method, called NoiseCutMix, to achieve natural, high-resolution image generation featuring the fused characteristics of two classes by partially combining the estimated noise corresponding to two different classes in a diffusion model. 
In the classification experiments, we verified the effectiveness of the proposed method by comparing it with conventional data augmentation techniques that combine multiple classes, random image generation using Stable Diffusion, and combinations of these methods.
Our codes are available at: \url{https://github.com/shumpei-takezaki/NoiseCutMix}.
\end{abstract}

\section{Introduction}
In recent years, data augmentation has been widely employed to improve the performance of deep learning~\cite{yang2022image,shorten2019survey,khalifa2022comprehensive}. 
Commonly used data augmentation techniques include methods that apply perturbations to a single image, such as adding noise, rotating, and scaling. 

\begin{figure}[t]
    \centering
    \includegraphics[width=.7\linewidth]{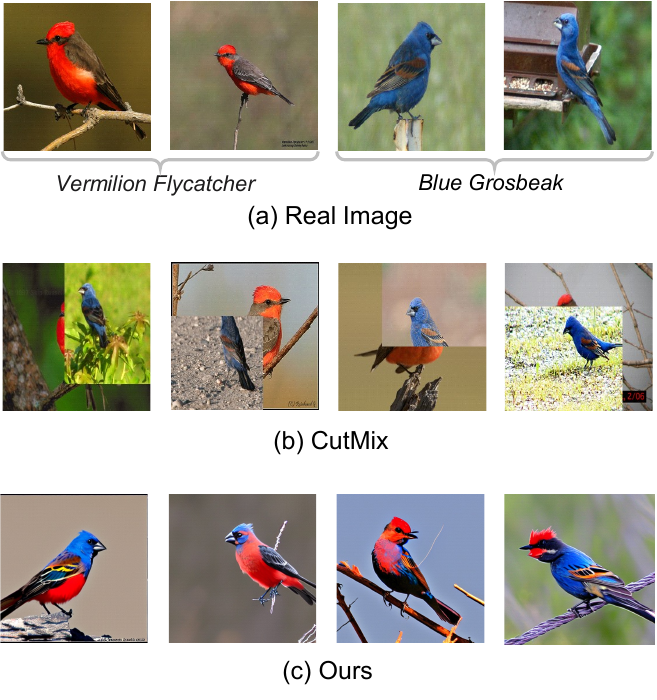}
    \vspace{-3mm}
    \caption{Comparison between images generated by CutMix and our data augmentation method.}
    \vspace{-2mm}
    \label{fig:overview}
\end{figure}

Moreover, techniques that combine images from two different classes, such as CutMix~\cite{yun2019cutmix} and MixUp~\cite{zhang2018mixup}, have also been proposed. 
CutMix augments data by cropping a random rectangle patch from one class image and pasting it onto another class image, yielding the mixed sample as shown in Figure~\ref{fig:overview}(b).
The augmented class label is obtained by weighting the two class labels in proportion to the patch area. Mixing images from different classes boosts data diversity and generalization. However, the unnatural boundaries between pasted regions can introduce structural inconsistencies that hinder feature learning.


A possible way to generate images that naturally merge the features of two source images is to use text-conditioned diffusion models~\cite{ho2020denoising} such as Stable Diffusion (SD)~\cite{Rombach_2022_CVPR}. These models can synthesize natural and high-resolution images in response to class labels or text prompts, and they readily produce scenes containing multiple objects. However, when the goal is to blend two classes, prompt engineering alone does not give fine-grained control over the relative contribution of each class, so achieving the precise class ratio required by CutMix remains difficult.

Hence, in this study, we propose NoiseCutMix, which integrates the CutMix idea into the generation process of a diffusion model. During each denoising step, we replace a spatial region of the estimated noise with the noise estimated for another class, using a binary mask whose area ratio directly controls how much each class contributes. This simple mechanism inherits the high-resolution synthesis ability of diffusion models while retaining the data-augmentation diversity of CutMix. As illustrated in the leftmost image of Figure~\ref{fig:overview}(c), our approach can generate a bird whose red breast resembles that of a ``Vermilion Flycatcher'' and whose blue head resembles that of a ``Blue Grosbeak.'' NoiseCutMix yields smoother class boundaries than conventional CutMix while allowing precise control over the mixing ratio.

In the classification experiments, we compared the proposed method with conventional data augmentation techniques that combine images from multiple classes, such as CutMix and MixUp, random image generation using Stable Diffusion, and the application of conventional data augmentation to the randomly generated images. We conducted evaluations using three datasets to verify the effectiveness of the proposed method. In particular, we confirmed that the images generated by our method effectively fuse features from different classes while reducing unnaturalness at the boundaries.

\section{Related Work}
Data augmentation by single-image perturbations, e.g., random rotations, rescaling, color shifts, or added noise, has been studied extensively \cite{yang2022image,shorten2019survey,khalifa2022comprehensive}.
These low-cost tricks help combat overfitting, but they change appearance rather than data structure, so the accuracy gains are modest.

Stronger techniques mix two images from different classes \cite{zhang2018mixup,yun2019cutmix}. MixUp~\cite{zhang2018mixup} linearly blends the two images, whereas CutMix~\cite{yun2019cutmix} pastes a rectangular patch from one onto the other. Although such cross-class mixing improves diversity and generalization, the pasted regions can look unnatural, and the resulting structural mismatch may hurt feature learning.

Recently, several data augmentation methods leveraging diffusion models have been proposed~\cite{zhou2024survey,alimisis2025advances}. 
Fine-tuning-based approaches~\cite{yuan2024realfake,trabucco2024effective,azizi2023synthetic} retrain Stable Diffusion on the data used to train the classifier, then use its generated images as data augmentation. Image-editing-based approaches~\cite{trabucco2024effective,Islam_2024_CVPR,Wang_2024_CVPR} make direct edits to real images using Stable Diffusion to enhance dataset diversity.
Methods that mix real and diffusion-generated images for a single class with fractal blending have been proposed~\cite{islam2024diffusemix,islam2024genmix}.

Our method, NoiseCutMix, enables the generation of natural and diverse images that fuse features from two different classes by leveraging the diverse data augmentation properties of CutMix within the generation process of Stable Diffusion.
Furthermore, our approach is effective yet simple, making it easy to plug into existing Stable Diffusion-based data augmentation methods.

\begin{figure}[t]
    \centering
    \includegraphics[width=.8\linewidth]{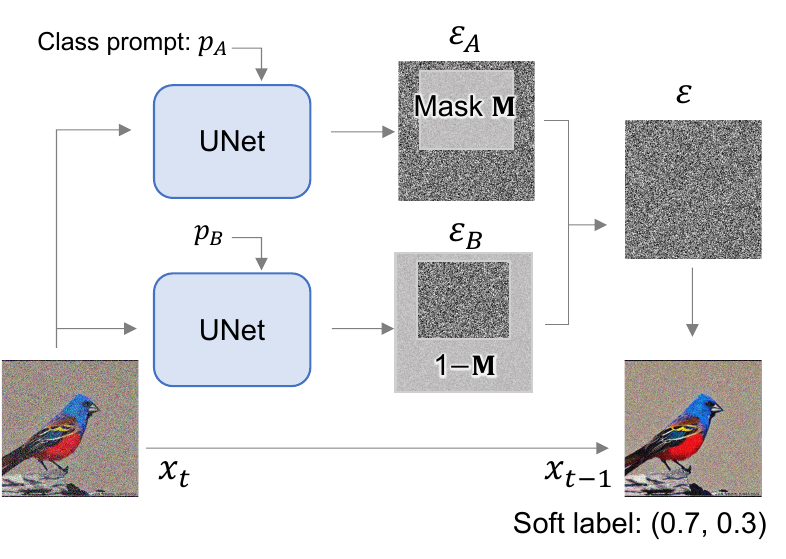}
    \vspace{-3mm}
    \caption{The proposed \emph{NoiseCutmix} mixes estimated noise of diffusion models in the denoising process. 
    }
    \vspace{-2mm}
    \label{fig:method}
\end{figure}

\section{NoiseCutMix: Data Augmentation by Mixing Estimated Noise in Diffusion Models}
We propose NoiseCutMix, a novel data augmentation method that leverages a pre-trained diffusion model, such as Stable Diffusion, to generate highly natural and diverse samples of a desired class, conditioned on text prompts.
As shown in Figure~\ref{fig:method}, NoiseCutmix blends the estimated noise from two different classes in a CutMix-like manner during the denoising process of a diffusion model. 
This fusion inherits two complementary strengths: (i) diffusion models generate high-resolution, natural images, while (ii) CutMix synthesizes diverse images by mixing features from different classes. Consequently, NoiseCutMix achieves augmented images that combine realism with variety.

\subsection{Mixing Estimated Noise of Diffusion Models}
The proposed method mixes the noise estimates of two class prompts during the reverse denoising process of a diffusion model to generate images that blend the visual traits of both classes. In a standard class-conditioned diffusion model, an image of a single target class is
obtained by iteratively removing noise from random Gaussian noise while conditioning the UNet on a text prompt $p$. 
In contrast, our method takes two class prompts, $p_A$ and $p_B$ corresponding to class $A$ and $B$, predicts their step-$t$ noises, and mixes them to generate an image that fuses the features of both classes.

Figure~\ref{fig:method} shows the overview of the proposed method at an arbitrary reverse step $t$.
Given the current image $x_t$, the Unet $f$ outputs two noise estimates, $\varepsilon_A = f(x_t, p_A, t)$ and $\varepsilon_B = f(x_t, p_B, t)$.
We then combine them with a binary mask $\mathbf{M} \in \{0,1\}^{W\times H}$ sampled uniformly at random over a rectangular region\footnote{$W$ and $H$ are the width and height of an image (or latent) features in a diffusion model}:
\begin{equation}
\varepsilon = \mathbf{M} \odot \varepsilon_A + (\mathbf{1}-\mathbf{M}) \odot \varepsilon_B,
\end{equation}
where $\odot$ denotes element-wise multiplication, and $\mathbf{1}$ is a binary mask of all ones. 

The mixed noise $\varepsilon$ is used to denoise the current image $x_t$ into $x_{t-1}$, which is fed to the next reverse step. Through this process, the final image inherits characteristics from both classes.
Details of how $\mathbf{M}$ is generated are described later.




\subsection{Class Labels for Generated Images}
Following the CutMix~\cite{yun2019cutmix}, the class label $\tilde{\mathbf{y}}$ (expressed in one-hot format) for the generated image is computed according to the region ratio $\lambda$ of the mask $\mathbf{M}$:
\begin{equation}
\tilde{\mathbf{y}} = \lambda \mathbf{y}_A + (1 - \lambda) \mathbf{y}_B,
\end{equation}
where $\lambda$ is sampled from a Beta distribution $\mathrm{Beta}(\alpha,\alpha)$ with a hyperparameter $\alpha$, as in CutMix. The vectors $\mathbf{y}_A$ and $\mathbf{y}_B$ denote the one-hot vector corresponding to class $A$ and $B$, respectively.

\subsection{Binary Mask for Mixing Estimated Noise}
We create the binary mask $\mathbf{M}$ for mixing estimated noise $\varepsilon_A$ and $\varepsilon_B$ by sampling a rectangular region with an aspect ratio proportional to the image (or latent) resolutions. The box coordinates of the mask are uniformly sampled according to a uniform distribution:
\begin{equation}
r_x \sim \mathrm{Unif}(0, W), \quad r_w = W \sqrt{1 - \lambda},
\end{equation}
\begin{equation}
r_y \sim \mathrm{Unif}(0, H), \quad r_h = H \sqrt{1 - \lambda}.
\end{equation}
Here, $(r_x, r_y, r_w, r_h)$ denotes the position and size of the rectangle, and the corresponding region is set to 0 in a mask $\mathbf{M}$. This sampling makes the cropped ratio $\frac{r_w r_h}{WH} = 1 - \lambda$.

\setlength{\tabcolsep}{12pt}
\begin{table*}[t]
\centering
\caption{Quantitative Evaluation: Accuracy [\%] on three datasets. We show the mean and standard deviation over five trials.}
\vspace{-2mm}
\label{tab:experiment}
\begin{tabular}{l|ccc}
\toprule
Method           & CUB~\cite{WahCUB_200_2011}       & Flower~\cite{nilsback2008automated}  & Aircraft~\cite{maji2013fine}       \\ \hline
Original     & \underline{67.78 (± 1.39)}  & 91.71 (± 1.09)  & 81.69 (± 0.90)   \\
CutMix~\cite{yun2019cutmix}      & 66.51 (± 1.20) & 89.93 (± 0.75) & {\bf 83.28 (± 0.32)}  \\
MixUp~\cite{zhang2018mixup}      & 67.57 (± 0.88) & 90.64 (± 0.52) & \underline{82.71 (± 0.84)} \\
SD-random~\cite{Rombach_2022_CVPR} & 67.60 (± 0.87) & \underline{92.52 (± 1.26)} & 78.27 (± 0.33)  \\
SD-random + CutMix & 62.90 (± 0.70) & 87.38 (± 1.01) & 77.78 (± 0.83)  \\
SD-random + MixUp & 64.57 (± 0.63) & 89.15 (± 0.72) & 78.36 (± 0.52)  \\ \hline
\rowcolor{gray!15} {\bf Ours}  & {\bf 68.78 (± 0.33)} & {\bf 92.91 (± 0.32)} & 79.69 (± 0.62)  \\ 
\bottomrule
\end{tabular}
\end{table*}
\begin{figure*}[t]
\centering
\includegraphics[width=.98\linewidth]{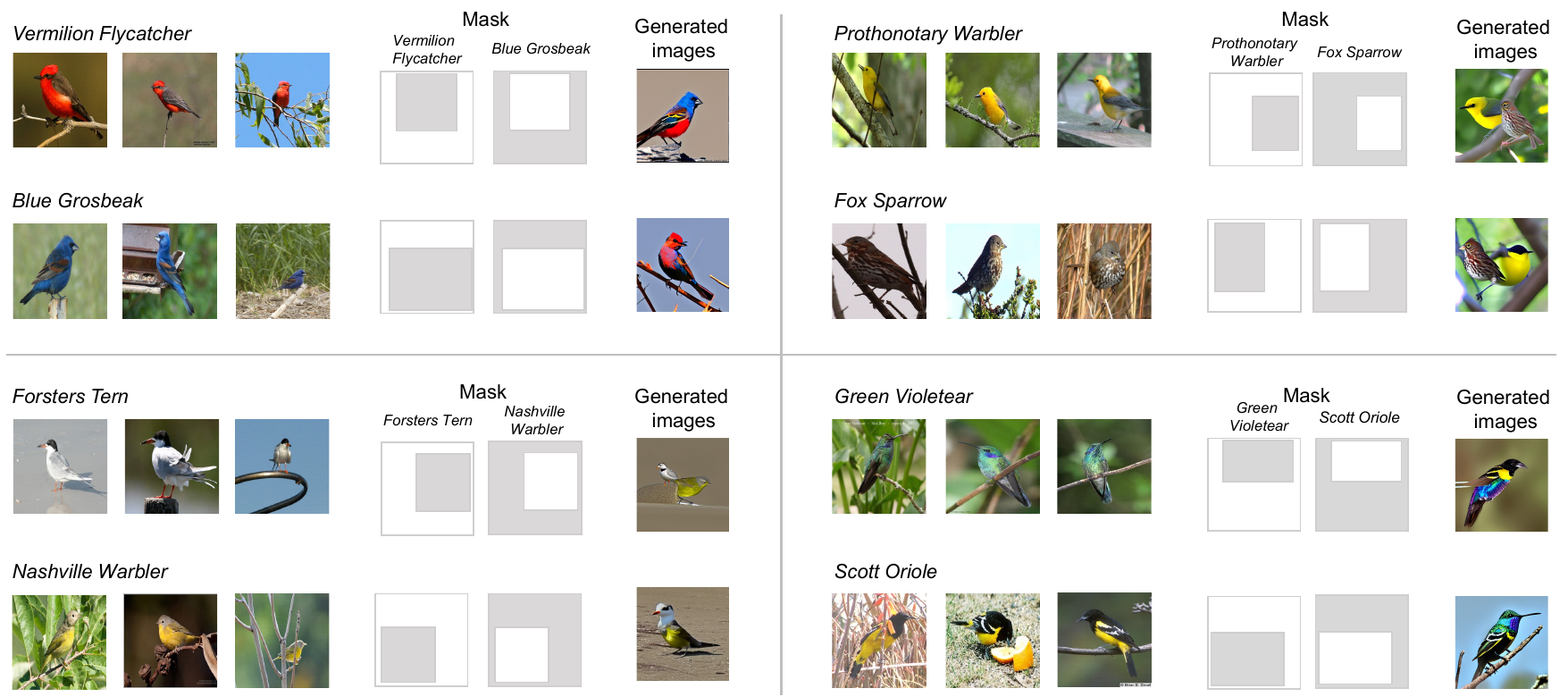}
\vspace{-2mm}
\caption{Examples from the CUB dataset: real images, images generated by our data augmentation method, and the masks used.}
\vspace{-2mm}
\label{fig:gen_img}
\end{figure*}
\section{Experiments}
\subsection{Comparison Methods}
To evaluate the effectiveness of the proposed method, we compared it with the following five methods.

\noindent
\textbf{CutMix~\cite{yun2019cutmix}:} A data augmentation method that replaces part of an image with a region from another image. Specifically, a rectangular region is sampled at random and cut out, then pasted onto the corresponding region of a different image, thus blending two images. The label is computed as $\tilde{\mathbf{y}} = \lambda \mathbf{y}_A + (1 - \lambda)\mathbf{y}_B$ based on the proportion of the cut region. We set the Beta distribution parameter $\alpha$ to 1.0 for sampling $\lambda$ and applied augmentation probability with 0.5.

\noindent
\textbf{MixUp~\cite{zhang2018mixup}:} A method that linearly interpolates two images at the pixel level. Specifically, two images are blended in proportion to $\lambda$, and the label is also interpolated as $\tilde{\mathbf{y}} = \lambda \mathbf{y}_A + (1 - \lambda)\mathbf{y}_B$. We set the Beta distribution parameter $\alpha$ to 0.2 for sampling $\lambda$ and also applied augmentation probability with 0.5.

\noindent
\textbf{SD-random:} Augment the dataset by randomly generating images with Stable Diffusion~\cite{Rombach_2022_CVPR}. Specifically, we used the text prompt ``a photo of a (\emph{class name})'' for class condition.

\noindent
\textbf{SD-random + CutMix/MixUp:} We took the images generated by SD-random above and applied CutMix or MixUp to them. In other words, this is a simple combination of Stable Diffusion augmentation with the conventional data augmentation methods.

\subsection{Datasets}
We used the following three datasets to evaluate our proposed method. During training, 20\% of the training images were randomly selected as validation data.

\noindent
\textbf{CUB~\cite{WahCUB_200_2011}:} Caltech-UCSD Birds (CUB) is a dataset for detailed bird classification, consisting of 11{,}788 images across 200 bird species. The split is 5{,}994 images for training and 5{,}794 for evaluation.

\noindent
\textbf{Flower~\cite{nilsback2008automated}:} Oxford Flowers is a dataset for detailed flower classification, including 8{,}189 images of 102 flower types. The split is 6{,}149 images for training and 2{,}040 for evaluation.

\noindent
\textbf{Aircraft~\cite{maji2013fine}:} FGVC-Aircraft is a dataset for detailed airplane classification, comprising 10{,}000 images of 102 airplane classes. The split is 6{,}667 images for training and 3{,}333 for evaluation.

\subsection{Implementation Details}
We used a ResNet50~\cite{he2016deep} pretrained on ImageNet as the classifier. The batch size was set to $64$, the learning rate to $0.001$, and we used Adam~\cite{kingma2015adam} as the optimizer. The number of epochs was set to $100$, and we adopted the parameters from the epoch with the highest accuracy on the validation set for evaluation.

Stable Diffusion v1.5 (SD) was employed as the diffusion model, and we used the text prompt ``a photo of (\emph{class name})'' for a class condition.  We also used Classifier-free guidance~\cite{ho2021classifierfree} with a guidance scale of $7.5$. Additionally, we employed DPMsolver++~\cite{NEURIPS2022_260a14ac} as the denoising sampler and set the inference steps to 25.

We performed fine-tuning for SD on each target dataset, following prior work~\cite{Wang_2024_CVPR}, which references LoRA~\cite{hu2021lora} and Textual Inversion~\cite{gal2023an}.\footnote{In practice, we used publicly available fine-tuned weights: \url{https://github.com/Zhicaiwww/Diff-Mix}} In our experiment, we generated the same number of augmented images as the original dataset (i.e., 100\% of the dataset size).

\section{Experimental Results}
\subsection{Quantitative Evaluation}
Table~\ref{tab:experiment} shows the classification results on the three datasets. “Original” refers to the result without any advanced data augmentation. We observe that our proposed data augmentation method outperforms conventional methods that combine multiple images (CutMix, MixUp) on CUB and Flower. In particular, we see improvements of 2.27\% on CUB and 2.98\% on Flower over CutMix. This is because our method mixes the estimated noise of two classes at each denoising step of Stable Diffusion, resulting in more natural boundaries between the combined images compared to CutMix. 
For Aircraft, neither NoiseCutMix nor SD-random improved accuracy because Stable Diffusion failed to match the dataset distribution, already degrading the original baseline. As NoiseCutMix depends on SD, no improvements were observed.

We also confirmed that our proposed method outperforms pure random image generation using Stable Diffusion (SD-random) on all datasets. This implies that simply generating random images with SD is less effective than merging multiple classes to increase data diversity. Moreover, our method also surpasses SD-random+CutMix, indicating that simply combining SD-generated images with conventional CutMix is insufficient; partial merging of noise estimates is crucial. \par

\subsection{Qualitative Evaluation}
Figure~\ref{fig:gen_img} shows real images, images generated by our proposed method, and the masks used, for four different class label pairs from the CUB dataset. For instance, the top-left example applies our data augmentation to classes “Vermilion Flycatcher” and “Blue Grosbeak.” 
These observations indicate that our method can generate images naturally blending features from two classes.

\section{Conclusion, Limitation, and Future Work}

NoiseCutMix blends estimated class-conditioned noise within diffusion models to synthesize images that coherently fuse two classes while preserving realism. Experiments show consistent gains over standard augmentation. 
A key limitation is dependence on Stable Diffusion (SD) aligning with the target data distribution, which can be mitigated by restricting training to in-distribution samples.
As future work, our approach, being both effective and simple, could be integrated into existing Stable Diffusion-based data augmentation pipelines.

\vspace{2mm}
\noindent
\textbf{{Acknowledgements:}} This work was supported by JSPS KAKENHI Grant Number JP23KJ1723, JP24KJ1805, and JP25K22846, and JST ACT-X Grant Number JPMJAX23CR, and ASPIRE Grant Number JPMJAP2403.

{
    \small
    \bibliographystyle{ieeenat_fullname}
    \bibliography{main}
}

\end{document}